\documentclass{article}

\usepackage[final,nonatbib]{nips_2016}

\usepackage[utf8]{inputenc} %
\usepackage[T1]{fontenc}    %
\usepackage{hyperref}       %
\usepackage{url}            %
\usepackage{booktabs}       %
\usepackage{amsfonts}       %
\usepackage{nicefrac}       %
\usepackage{microtype}      %

\usepackage{mathtools}
\usepackage{bm}
\usepackage{siunitx}
\usepackage{mathtools}
\usepackage{tikz}
\usepackage{pgfplots}
\usepackage{booktabs}
\usepackage{subfig}
\usepackage{xspace}
\usepackage[capitalize]{cleveref}
\usepackage[style=english,english=american]{csquotes}
\usepackage{enumitem}
\usepackage[autolanguage]{numprint}

\pgfplotsset{compat=newest}
\usetikzlibrary{plotmarks}
\usetikzlibrary{positioning}
\usetikzlibrary{decorations.markings}

\usepackage{xspace}
\usepackage{xparse}
\usepackage{mathtools}

\newcommand\dataset{\mathcal{D}}

\newcommand{\eg}{\emph{e.g.,}\xspace}

\newcommand\myparagraph[1]{\parindent0pt \vspace{8pt} \textbf{#1}\quad}
\renewcommand{\vec}[1]{\ensuremath{\mathbf{\boldsymbol{#1}}}}

\newcommand{\inputsingle}{\ensuremath{\vec{x}}}
\newcommand{\labelsymbol}{y}
\newcommand{\labelsingle}{\labelsymbol}

\newcommand{\noe}{\ensuremath{N}}

\newcommand{\expectation}{\ensuremath{\mathbb{E}}}

\newcommand{\modelfunction}{\ensuremath{f}}%

\newcommand{\loss}{\ensuremath{\mathcal{L}}}
\newcommand{\learningObjective}{\ensuremath{\bar{\loss}}}%

\newcommand{\regularizationfunction}{\ensuremath{\omega}}

\newcommand{\parameters}{\ensuremath{\vec{\theta}}}

\newcommand{\parametersUpdated}{\ensuremath{\parameters'}}

\newcommand{\inputAL}{\ensuremath{\inputsingle'}}
\newcommand{\inputlabelAL}{\ensuremath{\labelsingle'}}
\newcommand{\numLayers}{\ensuremath{L}}

\title{Active and Continuous Exploration with Deep Neural Networks and Expected Model Output Changes}

\author{
    Christoph K{\"a}ding$^{1,2}$\\
    \small\texttt{christoph.kaeding@uni-jena.de}\\
    \And
    Erik Rodner$^{1,2}$\\
    \small\texttt{erik.rodner@uni-jena.de}\\
    \And
    Alexander Freytag$^{3}$\\
    \small\texttt{alexander.freytag@zeiss.com}\\
    \And
    Joachim Denzler$^{1,2}$\\
    \small\texttt{joachim.denzler@uni-jena.de}\\
}

\begin{document}

\maketitle

\vspace{-1.5em}
{
    \centering
    $^1$Computer Vision Group, Friedrich Schiller University Jena, Germany\\
    $^2$Michael Stifel Center Jena, Germany\\
    $^3$Carl Zeiss AG, Jena, Germany\\
}

\ \\

\begin{abstract}
    The demands on visual recognition systems do not end with the complexity offered by current large-scale image datasets, such as ImageNet. %
    In consequence,
    we need curious and continuously learning algorithms that actively acquire knowledge about semantic concepts which are present in available unlabeled data.
    As a step towards this goal,
    we show how to perform continuous active learning and exploration,
    where an algorithm actively selects relevant batches of unlabeled examples for annotation.
    These examples could either belong to already known or to yet undiscovered classes.
    Our algorithm is based on a new generalization of the Expected Model Output Change principle for deep architectures and is especially tailored to deep neural networks.
    Furthermore,
    we show easy-to-implement approximations that yield efficient techniques for active selection.
    Empirical experiments show that our method outperforms currently used heuristics.
\end{abstract}

\section{Introduction}
Without any doubt,
supervised learning of visual recognition models has made an impressive progress in the last years~\cite{Agrawal14:APM,Babenko2014,Hariharan2014,Branson2014a,Girshick2014}.
This success is based on techniques which allow to learn complex and deep representations from well annotated and large-scale datasets.
However, to further foster the application of vision algorithms in related domains, such as biomedical image analysis or
quantification in data-driven science, we need to move from static algorithms which are trained only once to continuously learning algorithms.
Thereby,
given unlabeled data can be actively explored in the search
for relevant new training examples and even new visual concepts of completely unknown object classes.

Nonetheless, keeping the annotation effort at a minimum is crucial in most application domains. %
We are therefore proposing an active learning algorithm that allows for class discovery during the process of guided data annotation.
Our algorithm is based on the expected model output change (EMOC) principle originally presented
in \cite{Freytag14_SIE,Kaeding15_ALD}.
In this paper,
we show how to generalize it towards layered models, such as deep neural networks. Furthermore, we show
that simple back-propagation allows for easily approximating the EMOC criterion and for evaluating it on large-scale unlabeled datasets.

\myparagraph{Related Work}
The authors of \cite{wang2016cost} present an approach for active learning of deep neural networks based on several selection criteria.
A key ingredient for the proposed framework is the pseudo labeling of most certain samples and the update of network parameters based on those pseudo labels as well as user annotated samples.
In \cite{novotny16i-have} the question is studied how semantic parts of images can be shared between categories.
An active learning method based on the uncertainty if regions are containing useful parts is proposed.
An opposite selection criterion is used in \cite{krause2015unreasonable},
where most confident samples are selected for user annotation to deal with noisy web images.
The authors argue that even images with high confidence can often result in false decisions.
Hence, labeling these samples correctly could have a huge impact for further decisions of the model.
However,
although all presented active learning strategies are well motivated,
they miss the direct link to the ultimate goal of active learning: the reduction of future errors.
Furthermore,
these techniques only make use of the  \emph{current output} of a deep neural network which shall be trained over time.
In contrast,
our technique considers \emph{changes in all available parameterized layers} and implicitly combine them into a single criterion.
Note finally that active sampling of instances for mini-batch gradient descent as introduced in \cite{gao2015active} is closely related to active learning.
The difference is that active learning tries to reduce the labeling effort for sample acquisition whereas active sampling aims at reducing the number of gradient descent iterations when all labels are known.
The authors accelerate mini-batch gradient descent by guiding selection based on uncertainty, significance, and selection history of each sample.
The resulting approach is very similar to our criterion.

\section{Active Learning with Deep Neural Networks}

Continuously learning convolutional neural networks with given labeled datasets is just one step towards a lifelong learning pipeline for visual recognition.
In fact,
a similarly important aspect is the acquisition of reliable labels,
which is still costly when using human annotators for all available data.
Therefore,
we show how to perform active learning with convolutional neural networks,
where only informative, unlabeled examples are automatically selected for being labeled by a human expert.
In the following,
we first review the expected model output change principle which was originally developed by Freytag et al. \cite{Freytag14_SIE}
as a query strategy for active learning with Gaussian process regression models.
We then show that this principle can be applied to convolutional neural networks by utilizing the approximated gradients of the loss function.

\myparagraph{Preliminaries and Notation}
Let $\modelfunction(\inputsingle; \parameters)$ be the output of a neural network with parameters $\parameters$ for a given image $\inputsingle$.
Learning a network from a given labeled training set $\dataset = (\inputsingle_i, \labelsingle_i)_{i=1}^{\noe}$
boils down to minimizing a desired learning objective:
\begin{align}
    \label{eq:opt}
    \learningObjective(\parameters; \dataset)
       &=
    \frac{1}{\noe}
    \sum\limits_{i=1}^\noe
        \loss( \modelfunction( \inputsingle_i; \parameters), \labelsingle_i)
            +
        \regularizationfunction(\parameters)
    \enspace.
\end{align}
Common choices for the loss function $\loss$ are the quadratic loss for regression tasks or the softmax loss for multi-class classification scenarios.
The term $\regularizationfunction$ is usually an elastic-net regularization~\cite{zou2005regularization} that combines $L_2$ and $L_1$-regularization of the parameters $\parameters$.

In the following, we focus on layered models
$   \modelfunction( \inputsingle_i; \parameters)
      =
   \modelfunction_{\numLayers}
   \left(
      \ldots
      \left(
      \modelfunction_{2}
      \left(
          \modelfunction_{1} \left(  \inputsingle_i; \parameters_{1} \right);
          \parameters_{2}
      \right)
      \ldots
      \right);
   \parameters_{\numLayers}
   \right)$
with $\parameters = (\parameters_1, \ldots, \parameters_{\numLayers})$ denoting all parameters of the model.
Applying the chain rule allows for calculating the partial derivatives with respect to the parameters.
In combination with using a gradient descent optimization scheme this is usually referred to as
back-propagation~\cite{Rumelhart1986LRB}.

\myparagraph{Review of the Expected Model Output Change Principle}
Automatically choosing the \enquote{most beneficial} examples for labeling is a challenging task,
since we do not know the labels the human expert will assign to the examples at the time of selection.
Therefore,
the \enquote{usefulness} of the examples for model learning needs to be estimated by treating unknown labels as latent variables.
Most standard active learning strategies for approaching this goal are empirically motivated
(\eg selecting examples with highest classifier uncertainty,
smallest distance to the decision boundary, etc.).
In contrast,
the EMOC principle approximates the estimated reduction of errors to avoid redundant queries.
The redundancy of an example  $\inputAL$
with respect to the learning result can be measured by calculating the difference of outputs for models learned with and without the currently investigated example:
\begin{align}
    \label{eq:emoc}
    \triangle \modelfunction(\inputAL)
    &=
    \expectation_{\inputlabelAL | \inputAL} \,
    \expectation_{\inputsingle} \,\left(\,
       d (\modelfunction(\inputsingle; \parameters), \modelfunction(\inputsingle; \parametersUpdated))
    \,
    \right)
    \enspace.
\end{align}
Here,
$d(z, z') = \|z - z\|_1$ measures the difference between model outputs and
$\parametersUpdated$ are the model parameters obtained by additionally training with the labeled example $(\inputAL, \inputlabelAL)$.
Since the label $\inputlabelAL$ of the unlabeled example $\inputAL$ is unknown,
we need to marginalize over $\inputlabelAL$ in the above formula (first expectation operation).
Examples with a small value of $\triangle \modelfunction$
are likely not changing the model outputs when being annotated and added to the training set,
therefore, they are considered  as redundant.
In consequence,
active learning based on EMOC selects the example with the highest value of $\triangle \modelfunction$.

In practice,
the expectations in \cref{eq:emoc} are estimated with empirical means across the dataset (for $\expectation_{\inputsingle}$)
and predictive posteriors (for $\expectation_{\inputlabelAL|\inputAL}$) based on the results of the current model.

\myparagraph{EMOC for Layered Models and Deep Neural Networks}
The EMOC principle is defined irrespective of the specific type of model or learning algorithm.
However,
its naive implementation would require learning a model from scratch for each example $\inputAL$ of an unlabeled dataset.
Therefore, techniques are required for efficiently evaluating or approximating the model output of~$\modelfunction'$.
The authors of \cite{Freytag14_SIE} focus on Gaussian process regression, where a closed-form expression for $\triangle \modelfunction(\inputAL)$
can be derived, which is efficient to evaluate.

In our case, we could make use of the continuous learning strategy of warm-start optimization \cite{Kaeding16_FDN}.
However, this would still be too time consuming for larger sets of unlabeled data.
We are therefore taking a shortcut and using a stochastic gradient approximation with just a single sample to estimate model parameter updates:
\begin{equation}
   \label{eq:approx_model_change}
    \parametersUpdated - \parameters
      \approx
     \gamma \nabla_{\parameters} \learningObjective(\parameters; \dataset \cup {(\inputAL, \inputlabelAL)})
       \approx
    \gamma \nabla_{\parameters} \learningObjective (\parameters; (\inputAL, \inputlabelAL))
\end{equation}
for some constant $\gamma>0$.
We can further approximate the model output change by using a first-order approximation:
\begin{align}
\label{eq:emocapprox}
\| \modelfunction(\inputsingle; \parametersUpdated) - \modelfunction(\inputsingle; \parameters) \|_1
&\approx \| \nabla_{\parameters} \modelfunction(\inputsingle; \parameters)^T (\parametersUpdated - \parameters) \|_1
\overset{\eqref{eq:approx_model_change}}{\approx}
\gamma \| \nabla_{\parameters} \modelfunction(\inputsingle; \parameters)^T
\nabla_{\parameters} \learningObjective (\parameters; {(\inputAL, \inputlabelAL)}) \|_1 \enspace.
\end{align}
Please note that $\nabla_{\parameters} \modelfunction$ is indeed a Jacobian matrix that
can be easily calculated with back-propagation.

The approximation reveals an interesting relationship between the EMOC criterion
and continuously learning deep neural networks using gradient descent. If
$\| \nabla_{\parameters} \learningObjective (\parameters; (\inputsingle_i, \labelsingle_i)) \|_1$ is small,
the example $(\inputsingle_i, \labelsingle_i)$ will likely not lead to any significant change of the model
parameters during optimization and can therefore be considered as redundant.
The scalar product in Eq.~\eqref{eq:emocapprox}
is also reasonable since changes in some of the model parameters only lead to changes of the model output
if the respective component in $\nabla_{\parameters} \modelfunction(\inputsingle; \parameters)$ has a high
absolute value.
Hence, our method avoids labeling examples which are likely redundant.

Since, the marginalization over all possible labels $\inputlabelAL$ can be costly for an increasing number of classes,
we additionally use a maximum \textit{a-posteriori} approximation by only considering the most likely label $\hat{y}'$ inferred by the previous
model $\modelfunction$.
Furthermore, we consider active learning with sets of examples, where we are not selecting a single unlabeled example $\inputAL$
but a whole set $\bm{X}'$ of examples. This eases continuous learning~\cite{Kaeding16_FDN} and also allows for efficient labeling of
multiple examples at once. We assume that all examples in $\bm{X}'$  will obtain the same label $\inputlabelAL$. Our approximated
EMOC score for each set $\bm{X}'$ is then given by:
\begin{align}
    \tilde{\triangle}\modelfunction(\bm{X}') &= \sum_{\inputAL \in \bm{X}'}
    \expectation_{\inputsingle} \,
            \| \nabla_{\parameters} \modelfunction(\inputsingle; \parameters)^T
            \nabla_{\parameters} \learningObjective (\parameters; {(\inputAL, \hat{y}'(\bm{X}'))}) \|_1 \enspace.
\end{align}
Since an exhaustive optimization over all possible sets $\bm{X}'$ is infeasible, we are using $M$ random sets of fixed size $K$.

\section{Experiments}

In the following, we describe conducted experiments in detail including dataset, baseline methods, and required parameters.

\myparagraph{Network Architecture and Choice of Parameters}
Since our main interest is in image categorization and understanding,
we apply deep convolutional neural networks as model family.
In particular, we use a randomly initialized CIFAR10 baseline model \cite{krizhevsky2009learning} pretrained on initially known samples.
For initial training as well as for continuous fine-tuning, we choose a mini-batch size of \numprint{64} samples and a learning rate of $0.0001$.
The networks are trained with mini-batch gradient descent~\cite{ngiam2011optimization} using momentum~\cite{sutskever2013importance} of \numprint{0.9} and weight decay~\cite{Rumelhart1986LRB} of $0.0005$.
Following the strategy of \cite{Kaeding16_FDN}, we enforce every mini-batch during update process to contain a portion of old as well as novel samples.
Therefore, we weight old data with $\lambda = \numprint{0.9}$ and novel data with $1-\lambda = \numprint{0.1}$ during sample selection for mini-batch gradient descent.
Thus, we prevent over-fitting towards novel samples and incorporate a small fraction of novel data at each mini-batch gradient descent iteration.
To continuously incorporate new knowledge and data, we perform \numprint{1000} single mini-batch gradient descent iterations at each update step.

\myparagraph{Experimental Setup and Dataset}
To evaluate our approach, we use the CIFAR100 dataset \cite{krizhevsky2009learning}.
It contains small images ($32 \times 32$ pixel) of \numprint{100} classes with
\numprint{500} training and \numprint{100} test samples each.
We build an initial training set out of \numprint{10} randomly chosen classes with \numprint{100} samples.
The unlabeled pool consists of additional \numprint{100} samples of each known class as well as \numprint{100} samples of \numprint{10} randomly selected novel classes.
All available test samples of the \numprint{20} corresponding classes serve as test set.
The presented results on the test set are averaged over \numprint{9} initializations.
In each active learning step,
we select $M = \numprint{1000}$ random sets each comprised of $K = \numprint{25}$ samples.
During this investigation, each set contains only samples from a single class chosen randomly.
Each set is evaluated according to the different active learning algorithms under the objective of fast exploration of the unlabeled pool.
The selected batch is then labeled and incorporated incrementally into the deep neural network.

\newcommand\approach[1]{\textit{(#1)}}
\myparagraph{Baseline Methods}
The most trivial baseline is passive learning by random selection of one of the $M$ batches \approach{random}.
The authors of \cite{novotny16i-have} propose an active learning method selecting most uncertain examples.
We adapt this approach to our scenario by selecting sets with lowest average minimum class score \approach{min}
which is similar to the criterion used in \cite{wang2016cost,kapoor2010gaussian}.
A related approach is to prefer batches containing samples with lowest difference of the two highest class scores \cite{wang2016cost,joshi2009multi}.
For an extension to sample sets, we again use the mean score of all contained samples \approach{1-vs-2}.
In \cite{krause2015unreasonable}, an opposite strategy is proposed where only the most certain samples are selected for labeling \approach{max}.

\myparagraph{Results}
In \cref{fig:res}, it can be seen that all evaluated methods, except the max strategy of \cite{krause2015unreasonable},
provide higher gain of accuracy as well as faster discovery of unseen classes than mere random selection of batches.
The reason for the inferior performance of the max strategy, which has been shown to be valuable in noisy settings, is the selection of most certain samples which likely belong to already known classes.
Both, min \cite{novotny16i-have} and 1-vs-2 \cite{wang2016cost} perform almost equally.
We contribute this behavior to the properties of the dataset and claim that batches with lowest maximal classification scores can also not be assigned clearly to a single class.
Finally, we observe that our proposed method is able to outperform all considered baselines in almost all update steps in the presented scenario.
Hence, we draw the conclusion that active learning of deep neural networks with expected model output changes is possible and beneficial.
\newlength{\updatesizewidth}%
\newlength{\updatesizeheight}%
\setlength{\updatesizewidth}{.35\textwidth}%
\setlength{\updatesizeheight}{.13\textheight}%
\newlength{\XLabelDistupdatesize}%
\newlength{\YLabelDistupdatesize}%
\setlength{\XLabelDistupdatesize}{-3pt}%
\setlength{\YLabelDistupdatesize}{-3pt}%
\newlength{\updatesizeColSepLegend}
\setlength{\updatesizeColSepLegend}{6pt}%
\newlength{\lwupdatesize}
\newlength{\lwupdatesizeplot}
\setlength{\lwupdatesize}{1.0pt}%
\setlength{\lwupdatesizeplot}{2.5pt}%
\newlength{\mytikzplotColSepLegend}
\setlength{\mytikzplotColSepLegend}{8pt}
\begin{figure}[t]
    \centering
    \scriptsize
    \begin{minipage}[t]{0.85\textwidth}
        \centering%
        \vspace{0pt}%
        \definecolor{mycolor1}{rgb}{0.00000,0.00000,0.17241}%
\definecolor{mycolor2}{rgb}{1.00000,0.10345,0.72414}%
\begin{tikzpicture}

\begin{axis}[%
width=0.967346\updatesizewidth,
height=\updatesizeheight,
at={(0\updatesizewidth,0\updatesizeheight)},
scale only axis,
xmin=0,
xmax=2000,
xlabel={Number of added Samples},
ymin=28,
ymax=48,
ylabel={Average Accuracy [\%]},
axis x line*=bottom,
axis y line*=left,
xlabel shift=\XLabelDistupdatesize,
ylabel shift=\YLabelDistupdatesize
]
\addplot [color=blue,dashed,line width=2.0pt,forget plot]
  table[row sep=crcr]{%
0	28.5222222222222\\
25	29.9833333333333\\
50	30.7833333333333\\
75	30.9555555555556\\
100	31.4277777777778\\
125	31.5888888888889\\
150	32.3111111111111\\
175	33.15\\
200	34.1666666666667\\
225	34.5166666666667\\
250	34.6722222222222\\
275	34.9944444444444\\
300	35.1555555555556\\
325	35.5722222222222\\
350	35.8833333333333\\
375	36.0277777777778\\
400	36.25\\
425	36.5777777777778\\
450	36.7555555555556\\
475	36.9444444444444\\
500	37.1666666666667\\
525	37.2944444444444\\
550	37.5\\
575	37.8388888888889\\
600	38.25\\
625	38.5111111111111\\
650	38.75\\
675	38.8888888888889\\
700	39.0888888888889\\
725	39.4\\
750	39.4166666666667\\
775	39.7388888888889\\
800	39.9722222222222\\
825	40.1777777777778\\
850	40.5111111111111\\
875	40.8388888888889\\
900	40.7444444444444\\
925	40.9222222222222\\
950	41.0833333333333\\
975	41.3555555555556\\
1000	41.3444444444444\\
1025	41.6\\
1050	41.7444444444445\\
1075	41.5222222222222\\
1100	41.7777777777778\\
1125	42.1166666666667\\
1150	42.0944444444444\\
1175	42.7\\
1200	42.4777777777778\\
1225	42.8388888888889\\
1250	42.8722222222222\\
1275	42.9444444444444\\
1300	42.9555555555556\\
1325	43.0388888888889\\
1350	43.3055555555556\\
1375	43.7555555555555\\
1400	43.6722222222222\\
1425	43.7611111111111\\
1450	43.9722222222222\\
1475	43.9611111111111\\
1500	44.3666666666667\\
1525	44.3333333333333\\
1550	44.5888888888889\\
1575	44.65\\
1600	44.8777777777778\\
1625	44.6888888888889\\
1650	45.0222222222222\\
1675	44.8833333333333\\
1700	45.1\\
1725	45.7666666666667\\
1750	45.7055555555555\\
1775	45.85\\
1800	45.9166666666667\\
1825	46.0444444444445\\
1850	46.2111111111111\\
1875	46.2833333333333\\
1900	46.3944444444444\\
1925	46.6833333333333\\
1950	46.65\\
1975	46.3\\
2000	46.2444444444444\\
};
\addplot [color=green,dashed,line width=2.0pt,forget plot]
  table[row sep=crcr]{%
0	28.5222222222222\\
25	29.2555555555556\\
50	29.8833333333333\\
75	30.6722222222222\\
100	31.1555555555556\\
125	31.4944444444444\\
150	31.8333333333333\\
175	32.5944444444444\\
200	32.6944444444444\\
225	33.2888888888889\\
250	33.3944444444444\\
275	34.2333333333333\\
300	34.2944444444445\\
325	35.0888888888889\\
350	35.4166666666667\\
375	35.8722222222222\\
400	36.4833333333333\\
425	36.4555555555556\\
450	37.2333333333333\\
475	37.6\\
500	37.8666666666667\\
525	38.6333333333333\\
550	38.45\\
575	38.7833333333333\\
600	39.4111111111111\\
625	39.4055555555556\\
650	39.5722222222222\\
675	40.0666666666667\\
700	40.2888888888889\\
725	40.6\\
750	40.9722222222222\\
775	40.8611111111111\\
800	40.9833333333333\\
825	41.3055555555556\\
850	41.5333333333333\\
875	42.0055555555556\\
900	42.0722222222222\\
925	42.4666666666667\\
950	42.35\\
975	42.8055555555556\\
1000	42.8888888888889\\
1025	43.0388888888889\\
1050	43.3\\
1075	43.5555555555556\\
1100	43.8333333333333\\
1125	43.8055555555556\\
1150	43.9166666666667\\
1175	44.1222222222222\\
1200	44.2333333333333\\
1225	44.0777777777778\\
1250	44.2555555555556\\
1275	44.3277777777778\\
1300	44.5333333333333\\
1325	44.2611111111111\\
1350	44.4555555555555\\
1375	44.5666666666667\\
1400	44.8111111111111\\
1425	44.6555555555556\\
1450	44.7722222222222\\
1475	45.0166666666667\\
1500	45.0833333333333\\
1525	45.35\\
1550	45.2277777777778\\
1575	45.55\\
1600	45.5166666666667\\
1625	45.8166666666667\\
1650	45.9222222222222\\
1675	46.0055555555556\\
1700	46.0444444444445\\
1725	45.9222222222222\\
1750	46.0777777777778\\
1775	46.1944444444444\\
1800	46.3666666666667\\
1825	46.6166666666667\\
1850	46.6\\
1875	46.6722222222222\\
1900	46.7722222222222\\
1925	46.6\\
1950	46.7\\
1975	46.8722222222222\\
2000	46.7111111111111\\
};
\addplot [color=mycolor1,dashed,line width=2.0pt,forget plot]
  table[row sep=crcr]{%
0	28.5222222222222\\
25	29.2611111111111\\
50	29.9888888888889\\
75	30.4888888888889\\
100	31.1944444444444\\
125	31.5611111111111\\
150	32.15\\
175	32.1777777777778\\
200	32.6722222222222\\
225	33.0666666666667\\
250	33.4\\
275	33.7833333333333\\
300	34.3111111111111\\
325	34.4666666666667\\
350	35.1333333333333\\
375	35.1388888888889\\
400	36.1444444444444\\
425	36.6333333333333\\
450	36.9166666666667\\
475	37.4666666666667\\
500	37.9722222222222\\
525	38.3277777777778\\
550	38.9055555555556\\
575	39.3277777777778\\
600	39.4333333333333\\
625	40.2111111111111\\
650	40.3833333333333\\
675	40.7055555555555\\
700	40.7222222222222\\
725	41.0111111111111\\
750	41.4555555555556\\
775	41.4944444444444\\
800	41.65\\
825	41.7888888888889\\
850	41.6833333333333\\
875	41.9222222222222\\
900	42.35\\
925	42.2333333333333\\
950	42.6333333333333\\
975	42.9055555555556\\
1000	43.5055555555556\\
1025	43.4666666666667\\
1050	43.6388888888889\\
1075	43.7277777777778\\
1100	43.7888888888889\\
1125	44.0388888888889\\
1150	44.2\\
1175	44.0722222222222\\
1200	44.4111111111111\\
1225	44.6277777777778\\
1250	44.7333333333333\\
1275	44.7055555555556\\
1300	44.6944444444444\\
1325	44.8111111111111\\
1350	45.0444444444444\\
1375	44.9333333333333\\
1400	45.2888888888889\\
1425	44.9777777777778\\
1450	45.2055555555556\\
1475	45.3611111111111\\
1500	45.2944444444445\\
1525	45.3555555555555\\
1550	45.1\\
1575	45.4722222222222\\
1600	45.3277777777778\\
1625	45.6666666666667\\
1650	45.8222222222222\\
1675	45.7222222222222\\
1700	45.7111111111111\\
1725	45.95\\
1750	46.0055555555556\\
1775	46.2\\
1800	46.3722222222222\\
1825	46.1388888888889\\
1850	46.2222222222222\\
1875	46.1944444444444\\
1900	46.5611111111111\\
1925	46.3055555555555\\
1950	46.1\\
1975	46.2944444444444\\
2000	46.3222222222222\\
};
\addplot [color=mycolor2,dashed,line width=2.0pt,forget plot]
  table[row sep=crcr]{%
0	28.5222222222222\\
25	28.7388888888889\\
50	28.9055555555556\\
75	28.9055555555556\\
100	29.2888888888889\\
125	29.6888888888889\\
150	29.7222222222222\\
175	29.7277777777778\\
200	29.85\\
225	30.1055555555556\\
250	30.3\\
275	30.9055555555556\\
300	31.8388888888889\\
325	32.2777777777778\\
350	32.6166666666667\\
375	32.7555555555555\\
400	32.8611111111111\\
425	33.3722222222222\\
450	33.3388888888889\\
475	33.8388888888889\\
500	34.05\\
525	34.0722222222222\\
550	34.1888888888889\\
575	34.7055555555556\\
600	35.3611111111111\\
625	35.7888888888889\\
650	36.0222222222222\\
675	36.1666666666667\\
700	35.9722222222222\\
725	36.3666666666667\\
750	36.7277777777778\\
775	36.85\\
800	37.1611111111111\\
825	37.45\\
850	37.7777777777778\\
875	38.0111111111111\\
900	38.1333333333333\\
925	38.6166666666667\\
950	38.8888888888889\\
975	39.1222222222222\\
1000	39.15\\
1025	39.7333333333333\\
1050	40.1388888888889\\
1075	40.5666666666667\\
1100	40.5333333333333\\
1125	40.9222222222222\\
1150	41.3\\
1175	41.3055555555555\\
1200	41.7666666666667\\
1225	42.1666666666667\\
1250	42.3166666666667\\
1275	42.6166666666667\\
1300	42.8833333333333\\
1325	43.0555555555556\\
1350	43.1333333333333\\
1375	43.1222222222222\\
1400	43.3\\
1425	43.4888888888889\\
1450	43.6944444444444\\
1475	43.7\\
1500	43.8944444444444\\
1525	44.3555555555556\\
1550	44.4944444444445\\
1575	44.6944444444444\\
1600	44.7055555555556\\
1625	44.9944444444444\\
1650	45.1388888888889\\
1675	45.2055555555556\\
1700	45.2722222222222\\
1725	45.6333333333333\\
1750	45.6888888888889\\
1775	45.6722222222222\\
1800	45.9333333333333\\
1825	46.1055555555556\\
1850	46.0277777777778\\
1875	45.9222222222222\\
1900	46.2722222222222\\
1925	46.4055555555556\\
1950	46.65\\
1975	46.8666666666667\\
2000	46.9333333333333\\
};
\addplot [color=red,solid,line width=2.0pt,forget plot]
  table[row sep=crcr]{%
0	28.5222222222222\\
25	29.9555555555556\\
50	30.9444444444444\\
75	31.5611111111111\\
100	32.5333333333333\\
125	32.9722222222222\\
150	33.5166666666667\\
175	34.1777777777778\\
200	34.5555555555556\\
225	35.4833333333333\\
250	35.7055555555555\\
275	36.35\\
300	36.8111111111111\\
325	37.3\\
350	37.5944444444444\\
375	37.8722222222222\\
400	37.9111111111111\\
425	38\\
450	38.45\\
475	38.7111111111111\\
500	38.9055555555556\\
525	39.0055555555556\\
550	39.1833333333333\\
575	39.3611111111111\\
600	39.5944444444444\\
625	39.9833333333333\\
650	40.2277777777778\\
675	40.2888888888889\\
700	40.7333333333333\\
725	40.9833333333333\\
750	40.8722222222222\\
775	41.7722222222222\\
800	41.8944444444444\\
825	41.8944444444444\\
850	42.3722222222222\\
875	42.3944444444444\\
900	42.8111111111111\\
925	43.1111111111111\\
950	43.1722222222222\\
975	43.2555555555556\\
1000	43.45\\
1025	43.5777777777778\\
1050	43.5611111111111\\
1075	43.6388888888889\\
1100	44.0111111111111\\
1125	43.8833333333333\\
1150	44.2666666666667\\
1175	44.4555555555556\\
1200	44.3777777777778\\
1225	44.6111111111111\\
1250	44.6944444444444\\
1275	44.7277777777778\\
1300	45.05\\
1325	45.2444444444444\\
1350	45.1722222222222\\
1375	45.1777777777778\\
1400	45.2333333333333\\
1425	45.8277777777778\\
1450	45.7444444444444\\
1475	45.7944444444444\\
1500	45.9388888888889\\
1525	45.9444444444444\\
1550	45.8555555555556\\
1575	45.8222222222222\\
1600	46.1222222222222\\
1625	46.2611111111111\\
1650	46.3666666666667\\
1675	46.1055555555555\\
1700	46.3944444444444\\
1725	46.3555555555555\\
1750	46.3444444444444\\
1775	46.75\\
1800	46.7277777777778\\
1825	47.0944444444444\\
1850	47.1055555555556\\
1875	47.1944444444444\\
1900	47.1555555555556\\
1925	46.9833333333333\\
1950	47.0833333333333\\
1975	47.1444444444444\\
2000	47.2222222222222\\
};
\end{axis}
\end{tikzpicture}%
%
        \definecolor{mycolor1}{rgb}{0.00000,0.00000,0.17241}%
\definecolor{mycolor2}{rgb}{1.00000,0.10345,0.72414}%
\begin{tikzpicture}

\begin{axis}[%
width=0.967346\updatesizewidth,
height=\updatesizeheight,
at={(0\updatesizewidth,0\updatesizeheight)},
scale only axis,
xmin=0,
xmax=2000,
xlabel={Number of added Samples},
ymin=10,
ymax=21,
ylabel={\# Discovered Classes},
axis x line*=bottom,
axis y line*=left,
xlabel shift=\XLabelDistupdatesize,
ylabel shift=\YLabelDistupdatesize
]
\addplot [color=blue,dashed,line width=2.0pt,forget plot]
  table[row sep=crcr]{%
0	10\\
25	10.7777777777778\\
50	11.2222222222222\\
75	11.6666666666667\\
100	12\\
125	12.2222222222222\\
150	12.7777777777778\\
175	13.3333333333333\\
200	13.7777777777778\\
225	14.1111111111111\\
250	14.3333333333333\\
275	14.6666666666667\\
300	14.8888888888889\\
325	15\\
350	15.3333333333333\\
375	15.5555555555556\\
400	15.6666666666667\\
425	15.8888888888889\\
450	16.3333333333333\\
475	16.3333333333333\\
500	16.5555555555556\\
525	16.6666666666667\\
550	16.8888888888889\\
575	17.2222222222222\\
600	17.5555555555556\\
625	17.6666666666667\\
650	17.8888888888889\\
675	18\\
700	18.1111111111111\\
725	18.2222222222222\\
750	18.2222222222222\\
775	18.2222222222222\\
800	18.2222222222222\\
825	18.3333333333333\\
850	18.4444444444444\\
875	18.5555555555556\\
900	18.5555555555556\\
925	18.5555555555556\\
950	18.6666666666667\\
975	18.6666666666667\\
1000	18.6666666666667\\
1025	18.7777777777778\\
1050	18.7777777777778\\
1075	18.8888888888889\\
1100	19\\
1125	19\\
1150	19.2222222222222\\
1175	19.3333333333333\\
1200	19.4444444444444\\
1225	19.5555555555556\\
1250	19.6666666666667\\
1275	19.6666666666667\\
1300	19.6666666666667\\
1325	19.6666666666667\\
1350	19.6666666666667\\
1375	19.6666666666667\\
1400	19.6666666666667\\
1425	19.6666666666667\\
1450	19.6666666666667\\
1475	19.6666666666667\\
1500	19.7777777777778\\
1525	19.7777777777778\\
1550	19.7777777777778\\
1575	19.7777777777778\\
1600	19.7777777777778\\
1625	19.8888888888889\\
1650	20\\
1675	20\\
1700	20\\
1725	20\\
1750	20\\
1775	20\\
1800	20\\
1825	20\\
1850	20\\
1875	20\\
1900	20\\
1925	20\\
1950	20\\
1975	20\\
2000	20\\
};
\addplot [color=green,dashed,line width=2.0pt,forget plot]
  table[row sep=crcr]{%
0	10\\
25	10.7777777777778\\
50	11.5555555555556\\
75	12.2222222222222\\
100	12.7777777777778\\
125	13.3333333333333\\
150	13.6666666666667\\
175	14.1111111111111\\
200	14.7777777777778\\
225	14.8888888888889\\
250	15.3333333333333\\
275	15.6666666666667\\
300	15.7777777777778\\
325	16.1111111111111\\
350	16.4444444444444\\
375	16.6666666666667\\
400	16.8888888888889\\
425	16.8888888888889\\
450	17.5555555555556\\
475	17.8888888888889\\
500	18.2222222222222\\
525	18.4444444444444\\
550	18.4444444444444\\
575	18.8888888888889\\
600	19.3333333333333\\
625	19.3333333333333\\
650	19.3333333333333\\
675	19.3333333333333\\
700	19.3333333333333\\
725	19.3333333333333\\
750	19.3333333333333\\
775	19.3333333333333\\
800	19.3333333333333\\
825	19.4444444444444\\
850	19.5555555555556\\
875	19.7777777777778\\
900	19.8888888888889\\
925	19.8888888888889\\
950	19.8888888888889\\
975	19.8888888888889\\
1000	19.8888888888889\\
1025	20\\
1050	20\\
1075	20\\
1100	20\\
1125	20\\
1150	20\\
1175	20\\
1200	20\\
1225	20\\
1250	20\\
1275	20\\
1300	20\\
1325	20\\
1350	20\\
1375	20\\
1400	20\\
1425	20\\
1450	20\\
1475	20\\
1500	20\\
1525	20\\
1550	20\\
1575	20\\
1600	20\\
1625	20\\
1650	20\\
1675	20\\
1700	20\\
1725	20\\
1750	20\\
1775	20\\
1800	20\\
1825	20\\
1850	20\\
1875	20\\
1900	20\\
1925	20\\
1950	20\\
1975	20\\
2000	20\\
};
\addplot [color=mycolor1,dashed,line width=2.0pt,forget plot]
  table[row sep=crcr]{%
0	10\\
25	10.7777777777778\\
50	11.5555555555556\\
75	12.3333333333333\\
100	13\\
125	13.3333333333333\\
150	13.7777777777778\\
175	14.2222222222222\\
200	14.4444444444444\\
225	14.8888888888889\\
250	15.3333333333333\\
275	15.8888888888889\\
300	16.3333333333333\\
325	16.5555555555556\\
350	17\\
375	17.1111111111111\\
400	17.5555555555556\\
425	17.8888888888889\\
450	18.1111111111111\\
475	18.3333333333333\\
500	18.3333333333333\\
525	18.6666666666667\\
550	18.8888888888889\\
575	19\\
600	19.2222222222222\\
625	19.3333333333333\\
650	19.4444444444444\\
675	19.4444444444444\\
700	19.5555555555556\\
725	19.6666666666667\\
750	19.8888888888889\\
775	19.8888888888889\\
800	19.8888888888889\\
825	19.8888888888889\\
850	19.8888888888889\\
875	19.8888888888889\\
900	19.8888888888889\\
925	19.8888888888889\\
950	19.8888888888889\\
975	19.8888888888889\\
1000	20\\
1025	20\\
1050	20\\
1075	20\\
1100	20\\
1125	20\\
1150	20\\
1175	20\\
1200	20\\
1225	20\\
1250	20\\
1275	20\\
1300	20\\
1325	20\\
1350	20\\
1375	20\\
1400	20\\
1425	20\\
1450	20\\
1475	20\\
1500	20\\
1525	20\\
1550	20\\
1575	20\\
1600	20\\
1625	20\\
1650	20\\
1675	20\\
1700	20\\
1725	20\\
1750	20\\
1775	20\\
1800	20\\
1825	20\\
1850	20\\
1875	20\\
1900	20\\
1925	20\\
1950	20\\
1975	20\\
2000	20\\
};
\addplot [color=mycolor2,dashed,line width=2.0pt,forget plot]
  table[row sep=crcr]{%
0	10\\
25	10\\
50	10\\
75	10\\
100	10.1111111111111\\
125	10.2222222222222\\
150	10.2222222222222\\
175	10.2222222222222\\
200	10.2222222222222\\
225	10.3333333333333\\
250	10.3333333333333\\
275	10.5555555555556\\
300	11\\
325	11.2222222222222\\
350	11.4444444444444\\
375	11.4444444444444\\
400	11.4444444444444\\
425	11.6666666666667\\
450	11.6666666666667\\
475	12\\
500	12\\
525	12\\
550	12.2222222222222\\
575	12.5555555555556\\
600	13\\
625	13.2222222222222\\
650	13.5555555555556\\
675	13.5555555555556\\
700	13.7777777777778\\
725	13.8888888888889\\
750	14.3333333333333\\
775	14.5555555555556\\
800	14.8888888888889\\
825	15\\
850	15\\
875	15.4444444444444\\
900	15.5555555555556\\
925	15.7777777777778\\
950	16\\
975	16.3333333333333\\
1000	16.4444444444444\\
1025	16.6666666666667\\
1050	16.7777777777778\\
1075	17.1111111111111\\
1100	17.3333333333333\\
1125	17.4444444444444\\
1150	17.6666666666667\\
1175	17.7777777777778\\
1200	18.1111111111111\\
1225	18.3333333333333\\
1250	18.4444444444444\\
1275	18.6666666666667\\
1300	18.7777777777778\\
1325	18.8888888888889\\
1350	19.1111111111111\\
1375	19.2222222222222\\
1400	19.4444444444444\\
1425	19.5555555555556\\
1450	19.6666666666667\\
1475	19.6666666666667\\
1500	19.7777777777778\\
1525	19.7777777777778\\
1550	19.7777777777778\\
1575	19.8888888888889\\
1600	19.8888888888889\\
1625	19.8888888888889\\
1650	19.8888888888889\\
1675	20\\
1700	20\\
1725	20\\
1750	20\\
1775	20\\
1800	20\\
1825	20\\
1850	20\\
1875	20\\
1900	20\\
1925	20\\
1950	20\\
1975	20\\
2000	20\\
};
\addplot [color=red,solid,line width=2.0pt,forget plot]
  table[row sep=crcr]{%
0	10\\
25	11\\
50	11.6666666666667\\
75	12.2222222222222\\
100	12.8888888888889\\
125	13.3333333333333\\
150	13.8888888888889\\
175	14.7777777777778\\
200	15.3333333333333\\
225	16.1111111111111\\
250	16.3333333333333\\
275	16.7777777777778\\
300	17\\
325	17.4444444444444\\
350	17.7777777777778\\
375	17.8888888888889\\
400	17.8888888888889\\
425	18.1111111111111\\
450	18.2222222222222\\
475	18.3333333333333\\
500	18.4444444444444\\
525	18.5555555555556\\
550	18.7777777777778\\
575	19\\
600	19\\
625	19.2222222222222\\
650	19.4444444444444\\
675	19.4444444444444\\
700	19.5555555555556\\
725	19.7777777777778\\
750	19.7777777777778\\
775	20\\
800	20\\
825	20\\
850	20\\
875	20\\
900	20\\
925	20\\
950	20\\
975	20\\
1000	20\\
1025	20\\
1050	20\\
1075	20\\
1100	20\\
1125	20\\
1150	20\\
1175	20\\
1200	20\\
1225	20\\
1250	20\\
1275	20\\
1300	20\\
1325	20\\
1350	20\\
1375	20\\
1400	20\\
1425	20\\
1450	20\\
1475	20\\
1500	20\\
1525	20\\
1550	20\\
1575	20\\
1600	20\\
1625	20\\
1650	20\\
1675	20\\
1700	20\\
1725	20\\
1750	20\\
1775	20\\
1800	20\\
1825	20\\
1850	20\\
1875	20\\
1900	20\\
1925	20\\
1950	20\\
1975	20\\
2000	20\\
};
\end{axis}
\end{tikzpicture}%
    \end{minipage}
    \begin{minipage}[t]{0.14\textwidth}
        \vspace{0pt}
        \hfill\definecolor{mycolor1}{rgb}{0.00000,0.00000,0.17241}%
\definecolor{mycolor2}{rgb}{1.00000,0.10345,0.72414}%

\newenvironment{customlegend}[1][]{
    \begingroup
    \csname pgfplots@init@cleared@structures\endcsname
    \pgfplotsset{#1}
}{
    \csname pgfplots@createlegend\endcsname
    \endgroup
}

\def\addlegendimage{\csname pgfplots@addlegendimage\endcsname}%

\begin{tikzpicture}
    \begin{customlegend}[
            legend entries={
                random\\
                min~\cite{novotny16i-have}\\
                1-vs-2~\cite{wang2016cost}\\
                max~\cite{krause2015unreasonable}\\
                \textbf{ours}\\
            },
            legend cell align=left,
            legend columns=1,
            legend style={
                draw=none,
                /tikz/column 2/.style={column sep=\mytikzplotColSepLegend,},
                /tikz/column 4/.style={column sep=\mytikzplotColSepLegend,},
                /tikz/column 6/.style={column sep=\mytikzplotColSepLegend,},
                /tikz/column 8/.style={column sep=\mytikzplotColSepLegend,},
            },
        ]

        \addlegendimage{color=blue,dashed,line width=2.0pt}
        \addlegendimage{color=green,dashed,line width=2.0pt}
        \addlegendimage{color=mycolor1,dashed,line width=2.0pt}
        \addlegendimage{color=mycolor2,dashed,line width=2.0pt}
        \addlegendimage{color=red,solid,line width=2.0pt}
    \end{customlegend}
\end{tikzpicture}
    \end{minipage}
    \caption{
        Comparison of different active learning approaches on the CIFAR100 dataset.
    }
    \label{fig:res}
\end{figure}
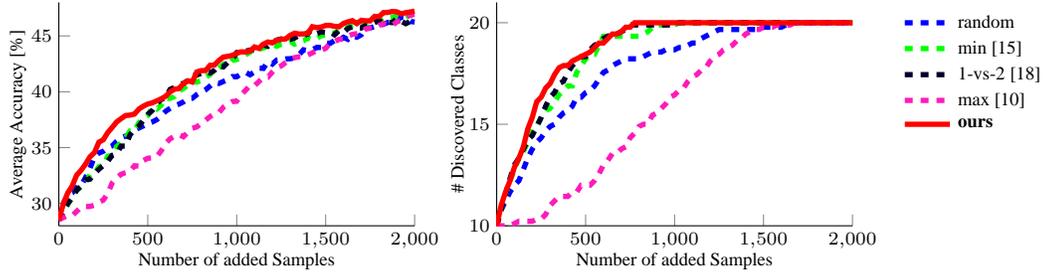

\section{Conclusions}
We presented an active learning algorithm based on the expected model output change principle,
which we extended to deep neural networks.
Proof-of-concept experiments on CIFAR100 using ground-truth labels for data batch creation promise good performance of the proposed idea.
A deeper evaluation with more general scenarios, other datasets, and deeper networks is left as future work.

{
    \scriptsize
    \myparagraph{Acknowledgements}
    This research was supported by grant DE 735/10-1 of the German Research Foundation (DFG).
}

\bibliographystyle{plain}
\bibliography{paper}

\end{document}